\documentclass[conference]{IEEEtran}
\IEEEoverridecommandlockouts
\usepackage{amssymb}
\usepackage{amsmath}
\usepackage{multicol}
\usepackage{algorithm}
\usepackage{algpseudocode}

\usepackage{multirow}
\usepackage{threeparttable}
\usepackage{graphicx}
\usepackage{balance}
\usepackage{array}
\usepackage{url}
\def\BibTeX{{\rm B\kern-.05em{\sc i\kern-.025em b}\kern-.08em
    T\kern-.1667em\lower.7ex\hbox{E}\kern-.125emX}}
\begin{document}

\title{State-of-the-art Advances of Deep-learning Linguistic Steganalysis Research\\
\thanks{This paper is accepted by \textit{2023 International Conference on Data, Information and Computing Science.}} 
\thanks{This work is supported by the National Natural Science Foundation of China (Grant U21B2020), supported by the Fundamental Research Funds for the Central Universities (Beijing university of posts and telecommunications) for Action Plan (2021XD-A11-1), and supported by BUPT Excellent Ph.D. Students Foundation (Grant CX2023120)}
}

\author{\IEEEauthorblockN{Yihao Wang}
	\IEEEauthorblockA{School of Cyberspace Security,\\
		Beijing University of Posts and Telecommunications,\\
		Email: \textit{yh-wang@bupt.edu.cn}}
	\IEEEauthorblockN{Yifan Tang}
	\IEEEauthorblockA{Beijing University of Posts and Telecommunications,\\
	Beijing, China.\\
Email: \textit{tyfcs@bupt.edu.cn}}
	\and
	\IEEEauthorblockN{Ru Zhang*}
	\IEEEauthorblockA{School of Cyberspace Security,\\
		Beijing University of Posts and Telecommunications,\\
		Email: \textit{zhangru@bupt.edu.cn}}
	\IEEEauthorblockN{Jianyi Liu}
	\IEEEauthorblockA{Beijing University of Posts and Telecommunications,\\
		Beijing, China.\\
	Email: \textit{liujy@bupt.edu.cn}}
}

\maketitle

\begin{abstract}
With the evolution of generative linguistic steganography techniques, conventional steganalysis falls short in robustly quantifying the alterations induced by steganography, thereby complicating detection. Consequently, the research paradigm has pivoted towards deep-learning-based linguistic steganalysis. This study offers a comprehensive review of existing contributions and evaluates prevailing developmental trajectories. Specifically, we first provided a formalized exposition of the general formulas for linguistic steganalysis, while comparing the differences between this field and the domain of text classification. Subsequently, we classified the existing work into two levels based on vector space mapping and feature extraction models, thereby comparing the research motivations, model advantages, and other details. A comparative analysis of the experiments is conducted to assess the performances. Finally, the challenges faced by this field are discussed, and several directions for future development and key issues that urgently need to be addressed are proposed.
\end{abstract}


\section{Introduction}
Steganography allows hidden communication by embedding information in digital media, such as texts [1]. The rise of big data and widespread use of platforms like Twitter and Facebook offer new opportunities for these concealed messages, making detection harder. The boom in social platforms enriches linguistic steganography, with advances from traditional methods to high-quality generative methods [2-4]. The misuse of steganography poses significant global threats, highlighting the importance of detecting concealed content amidst vast online data.

Linguistic steganalysis has been developed to detect the existence of concealed information within textual content, acting as a protective measure against steganographic techniques [5]. To date, researchers have proposed a variety of linguistic steganalysis works [6-34]. These works can be broadly stratified into two categories based on their research focus and feature extraction methodologies: traditional linguistic steganalysis [5,6] and deep-learning linguistic steganalysis [7-34]. Traditional methods primarily focus on constructing effective word associations and other manually engineered statistical features for detection. However, due to the high statistical concealment achieved by generative steganography in generating stego texts [2-4], traditional methods suffer from limited feature richness, insufficient robustness in quantifying steganographic perturbations, resulting in lower performance and poor generalization. Consequently, the research emphasis has shifted towards designing deep-learning steganalysis models. Deep learning steganalysis captures the changes in word correlations caused by steganography embeddings and extracts highly diverse features, leading to improvements in steganalysis [7-9].

The remaining sections of this paper are organized as follows: Section \ref{sec2} provides a overview of linguistic steganalysis. Section \ref{sec3} categorizes existing work and analyzes their model details. Section \ref{sec4} presents an analysis of the experimental aspects of existing deep-learning methods. Section \ref{sec5} discusses the challenges that need to be addressed.

\section{Preliminaries}\label{sec2}
Existing deep-learning models for linguistic steganalysis have the capability to extract diverse steganalysis features. In this study, we formalize deep-learning linguistic steganalysis with the following expression.

\begin{equation}\label{eq1}
P = {\text{Classifier}}({\mathbf{F}}),
\end{equation}

\begin{equation}\label{eq2}
{\mathbf{F}} = {\text{TStega - Net}}(V),
\end{equation}

\begin{equation}\label{eq3}
V = {\text{Embed}}(T),
\end{equation}

\begin{equation}\label{eq4}
T = C \cup S = \{ x_1^c, \cdots ,x_n^c,x_1^s, \cdots ,x_m^s\}.
\end{equation}

\begin{figure*}[!htbp]
	\begin{center}
		\includegraphics[height=2.6cm, width=17.5cm]{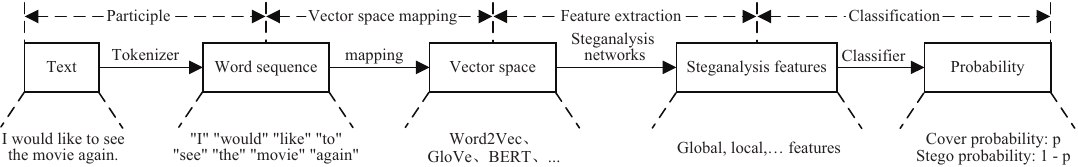}
	\end{center}
	\caption{Overall workflow of deep-learning linguistic steganalysis models.}
	\label{fig1}
\end{figure*}

\noindent where, Classifier(·) represents the classifier, which performs the detection task using the extracted features. TStega-Net(·) represents the deep linguistic steganalysis models that have been designed. Embed(·) represents to the text vectorization model that maps the texts into a specific vector space. $T$ is a given text set, consisting of a set $C$ of $n$ cover texts $x_i^c$ and a set $S$ of $m$ stego texts $x_i^s$. $V$ is the vector space in which the text set is mapped. $P$ is the probabilities of the cover texts and stego texts. $\mathbf{F}$ is the features extracted by the steganalysis network, encompassing various aspects such as text semantics, including inter-word correlations [7,8,11,12,16,20], and text structure, including sentence-level correlations \cite{33} as well as syntax and grammar-related features [14,32]. The overall workflow of deep-learning linguistic steganalysis models is visualized in Figure \ref{fig1}.

\section{Design details of existing work}\label{sec3}
\subsection{Overview}\label{sec31}
\begin{figure}[!htbp]
	\begin{center}
		\includegraphics[height=11cm, width=8cm]{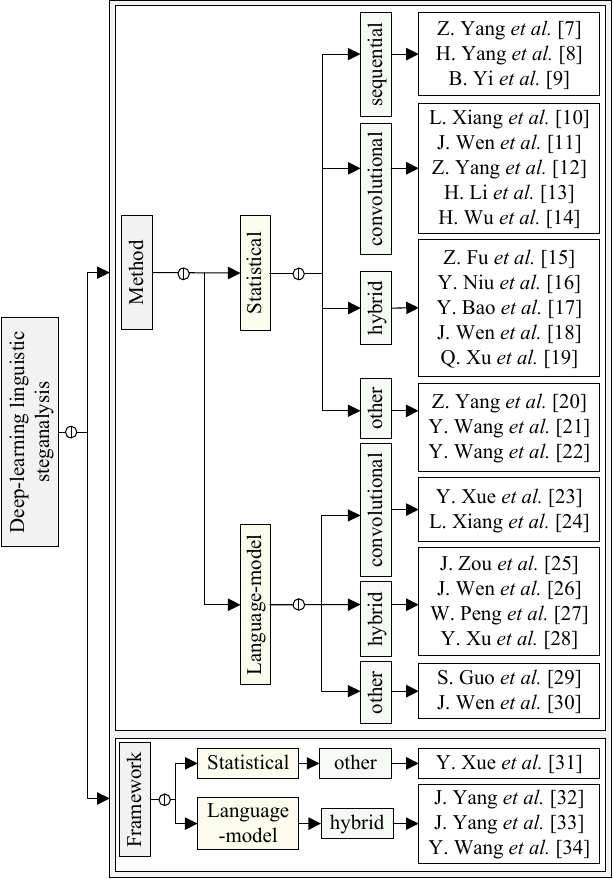}
	\end{center}
	\caption{Classification of deep-learning linguistic steganalysis.}
	\label{fig2}
\end{figure}

Overall, deep-learning linguistic steganalysis involves mapping the text to a vector space, feeding the vector representation into a designed deep steganalysis network, extracting features that exhibit distribution changes caused by steganography, and using features to determine whether the text is stego. The classification is shown as Figure \ref{fig2}.

\textbf{Primary classification}: Based on the distinct in vector space mapping. Existing work can be classified into two categories: "statistical vector embedding" [7-22,31] and "language-model vector embedding" [23-30,32-34]. The "statistical vector embedding" methods offer advantages such as simplicity in implementation and shorter training time. Researchers have utilized pre-trained models like Word2Vec \cite{35} and GloVe \cite{36} to achieve steganalysis models with improved performance compared to traditional methods. With the emergence of large-scale language models \cite{37}, linguistic steganalysis has also started incorporating these models as semantic mapping tools, which we refer to as "language-model vector embedding." This type of work benefits from higher-dimensional semantic spaces, leading to further improvements in performance.

\textbf{Secondary classification}: Based on the distinct feature extraction models. Existing work can be further categorized into four classes: "sequential," "convolutional," "hybrid," and "other." The "sequential" methods solely employ architectures based on RNN (Recurrent Neural Network), LSTM (Long Short-Term Memory), and other sequential models. The "convolutional" methods solely employ architectures based on CNN (Convolutional Neural Network), GCN (Graph Convolution Network), and other convolutional models. The "hybrid" methods combine different types of network architectures, such as the combination of LSTM and CNN \cite{16}. The "other" methods use alternative models.

\subsection{"Statistical vector embedding" method details}\label{sec32}
\begin{table*}[!htbp]
	\centering	
	\scriptsize
	\caption{Model details of “statistical vector embedding” linguistic steganalysis methods.}
	\label{tab1}
	\begin{tabular}{c|c|c|cc|c}
		\hline
		\textbf{Ref.} & \textbf{Research motivation} & \textbf{Model advantages} & \multicolumn{2}{c|}{\textbf{Basic formulas}} & \textbf{Category} \\ \hline
		\cite{7} & \multirow{3}{*}{\shortstack[c]{Capturing long-distance correlations\\ between words to improve the \\performance of steganalysis.}}
		& \multirow{3}{*}{\shortstack[c]{Extracting dependencies between \\ words at longer distances.}} & 
		${C_t} = f{o_t} \odot {C_{t - 1}} + {i_t} \odot \tanh ({\widetilde C_t})$ & & \multirow{3}{*}{sequential} \\ \cline{1-1}
		\cite{8} & ~ & ~ & ${O_t} = \sigma ({W_o} \times [h{i_{t - 1}},{x_t}] + {b_o})$  &(5) & ~\\ \cline{1-1}
		\cite{9} & ~ & ~ & $h{i_t} = {O_t} \odot \tanh ({C_t})$ & & ~\\ \hline
		\cite{10} & \multirow{5}{*}{\shortstack[c]{Capturing strong associations \\of surrounding words to\\ improve steganalysis performance.}} & \multirow{3}{*}{\shortstack[c]{Capturing strong correlations between \\ words at short distances.}} &${{\mathbf{F}}^{n - 1}} = \sum\nolimits_i {{\mathbf{F}}_i^{l - 1} * K_i^l + b_i^l}$ & \multirow{2}{*}{(6)} & \multirow{8}{*}{convolutional} \\ \cline{1-1}
		\cite{11} & ~ & ~ & ${{\mathbf{F}}^l} = {\text{pool}}({f^n}({{\mathbf{F}}^{n - 1}}))$ & & ~ \\ \cline{1-1}
		\cite{12} & ~ & ~ & ~ & & ~ \\ \cline{1-1}\cline{3-5}
		\cite{13} & ~ & \shortstack[c]{Capturing strong correlations between \\ words at short distances.} &${\alpha _{ij}} = \frac{{\exp ({a^{\text{T}}}[W{x_i}||W{x_j}])}}{{\sum\nolimits_{k \in {N_i}} {\exp ({a^{\text{T}}}[W{x_i}||W{x_j}])} }}$
		& \multirow{3}{*}{(7)}& ~ \\ \cline{1-3}
		\cite{14} & \shortstack[c]{Using a globally-shared matrix to extract\\ correlations, improving performance.} & Having a strong ability to capture grammar. & $x_i^M = \sigma (\frac{1}{K}\sum\limits_{k = 1}^K {\sum\limits_{j \in {N_i}} {\alpha _{ij}^k{W^k}x_j^{M - 1}}})$ & & ~ \\ \hline
		\cite{15} & \multirow{5}{*}{\shortstack[c]{Combining the advantages of \\different networks to improve the\\ detection performance of stego texts.}} & \multirow{4}{*}{\shortstack[c]{Incorporating the advantages of \\ different network models to capture \\ diverse types of features.}} & \multicolumn{2}{c|}{Eq. (5) Eq. (7)} & \multirow{5}{*}{hybrid} \\ \cline{1-1}\cline{4-5}
		\cite{16} & ~ & ~ & \multicolumn{2}{c|}{\multirow{4}{*}{Eq. (5) Eq. (6)}} & ~ \\ \cline{1-1}
		\cite{17} & ~ & ~ &  & & ~ \\ \cline{1-1}
		\cite{18} & ~ & ~ & ~ & & ~ \\ \cline{1-1}\cline{3-3}
		\cite{19} & ~ & Considering the significance of features. & ~ & & ~ \\ \hline
		\cite{20} & \shortstack[c]{Improving the performance \\ of linguistic steganalysis.} & Implementing a simple and fast approach. & ${h^{(k)}} = \frac{1}{{{L^{(k)}}}}{\sum\nolimits_{i = 1}^{{L^{(k)}}} {A[{x_i}D]} ^{\rm T}}$& (8)& other \\ \hline
	\end{tabular}
\end{table*}

Existing work on "statistical vector embedding" includes: 

\textbf{1. Global Feature Extraction and Long-term Dependencies}: Yang \textit{et al.} \cite{7} employed an RNN to extract the conditional probability distribution of words, capturing global features to obtain longer word dependencies. Yang \textit{et al.} \cite{8} proposed a method based on feature pyramid dense connections, which captures dependencies at different levels and achieves superior detection performance. Yi \textit{et al.} \cite{9} introduced two novel models by pre-training and fine-tuning steganalysis classifiers: one based on pre-trained RNN language models and the other based on pre-trained sequence autoencoder models.

\textbf{2. Local Feature Extraction and Short-range Dependencies}: Xiang \textit{et al.} \cite{10} designed a CNN-based steganalysis method that effectively detects stegos generated by modified steganography schemes. Wen \textit{et al.} and Yang \textit{et al.} \cite{11,12} proposed methods based on different CNN structures. Li \textit{et al.} \cite{13} designed a steganalysis capsule network that analyzes the subtle differences between stego texts and cover texts at low payloads. Wu \textit{et al.} \cite{14} applied GCN to steganalysis, transforming text into a directed graph with relevant information and updating the nodes in the graph by collecting contextual information to enhance self-expression of each word.

\textbf{3. Hybrid Architectures and Enhanced Feature Extraction}: To leverage the advantages of different architectures, Fu \textit{et al.} \cite{15} fused features from different levels. Niu \textit{et al.} \cite{16}, Bao \textit{et al.} \cite{17}, and Wen \textit{et al.} \cite{18} combined CNN and LSTM in different ways, capturing global and local information and extracting multi-granularity features. Considering that the importance of different features may vary, Xu \textit{et al.} \cite{19} employed a mechanism of group enhancement to enhance the important features. 

\textbf{4. Other Architectures}: Yang \textit{et al.} \cite{20} proposed a fast and effective method that extracted correlations between words, achieving significant improvements in performance compared to traditional methods. Wang \textit{et al.} \cite{21} designed a method with a variable-scale network. Depending on the text length, the method controlled the network's scale using a parameter setting mechanism, achieving a reduction in model size and training time while improving performances. Wang \textit{et al.} \cite{22} provides an autonomous solution for steganalysis that uses reinforcement learning to effectively detecting stego texts in distribution-tranformed scenarios.

We conduct a examination of the specifics of "statistical vector embedding" methods, as outlined in Table \ref{tab1}.

In Eq.(5), $hi_t$ represents the hidden state of LSTM at the $t$th step, $fo_t$, $i_t$, and $O_t$ separately represent the forget, input, and output gates at the $t$th step, and $C_t$ represents the memory cell at time step $t$. The activation function is denoted by $\sigma(\cdot)$. In Eq.(6), ${\mathbf{F}}^l$ represents the features extracted by the $l$th CNN layer, $K^l$ and $b^l$ represent the convolutional kernel and bias of the $l$th CNN layer, and $pool(\cdot)$ represents the pooling operation. In Eq.(7), $\alpha _{ij}$ represents the attention weight of the edge connecting nodes $i$ and $j$, $N$ represents the set of neighboring nodes of $i$, and $K$ represents the number of attention mechanisms. In Eq.(8), $L^{(k)}$ represents the $k$th word in a sentence,   contains the extracted word semantic-related features, $D$ is the dictionary, $A$ represents the word correlation matrix used to extract semantic relatedness between words in the input sentence, $x$ represents the input of the node, and $\operatorname{T}$ represents the transpose operation.

\subsection{"Language-model vector embedding" method details}\label{sec33}
Existing work on "language-model vector embedding" includes:

\textbf{1. Local Feature Extraction and Short-range Dependencies}: Xue \textit{et al.} \cite{23} proposed a domain adaptive linguistic steganalysis method to alleviate the issue of low detection performance caused by domain mismatch. They also designed a distributed adaptive layer and employed three loss functions. Xiang \textit{et al.} \cite{24} used GAT to extract global features, ensuring collaboration between BERT's local features and GAT's global features in the joint prediction layer.

\textbf{2. Hybrid Architectures and Enhanced Feature Extraction}: Zou \textit{et al.} \cite{25} proposed an outstanding steganalysis method, which maps words to the semantic space of BERT and using bidirectional LSTM and attention mechanisms to extract locally critical features. To enable a small number of samples to train the network and enhance generalization to different detection tasks, Wen \textit{et al.} \cite{26} introduced a meta-learning method. Peng \textit{et al.} \cite{27} introduced transfer learning to improve the performance of LSTM and CNN-based methods by leveraging BERT's guidance in word vector representation. Xu \textit{et al.} \cite{28} explored the interaction between local and global features in texts. 

\textbf{3. Other Architectures}: Guo \textit{et al.} \cite{29} encoded TF-IDF statistical features to effectively combine semantic and statistical features. Wen \textit{et al.} \cite{30} proposed a novel method. It created a task sequence arranged in chronological order and introduced the concept of lifelong learning.

\subsection{Linguistic steganalysis framework details}\label{sec34}
The introduction of frameworks requires extracting motivations into more universal frameworks that can improve the performance of existing methods in specific scenarios. Xue \textit{et al.} \cite{31} proposed an alternative hierarchical co-learning steganalysis framework, which alleviates the challenges of large-scale models. Yang \textit{et al.} \cite{32} designed a novel framework that preserves the semantic information of the text and fully uses the syntactic structure. Yang \textit{et al.} \cite{33} constructed a steganalysis framework specifically targeting social networks. This framework uses existing methods and an aggregated GNN (Graph Neural Network) as the content embedding and context embedding. Wang \textit{et al.} \cite{34} built a framework by the user profile for enhancing steganalysis performance. This framework can not only improve the performance of existing steganalysis methods in social networks, but also enable related-task methods to exert their potential effectiveness in steganalysis tasks.

\section{Experiments}\label{sec4}
\subsection{Results and analysis}\label{sec42}

\begin{table}[!htbp]
	\centering
	\setlength{\tabcolsep}{1.65mm}
	\scriptsize
	\begin{threeparttable}
		\caption{Acc comparison of different methods in traditional laboratory data. (Stegenography scheme: RNN-Stega (FLC and VLC) \cite{3})}
		\label{tab2}
		\begin{tabular}{c|c|c|ccc|ccc}
			\hline
			\multicolumn{3}{c|}{Dataset} & \multicolumn{6}{c}{\textbf{Twitter}}\\ 
			\hline
			\multicolumn{3}{c|}{bpw} & \multicolumn{3}{c|}{FLC} & \multicolumn{3}{c}{VLC}\\
			\cline{1-3}
			\multicolumn{2}{c|}{classification} & method & 1&2&3&1&2&3 \\ 
			\hline
			\multirow{9}{*}{S}  & \multirow{3}{*}{sequential} & [7] & 0.900&	0.882&	0.895&	0.904&	0.881&	0.887 \\ 
			~ & ~ & [8] &$\backslash$&$\backslash$&$\backslash$&0.929&0.923& 0.887\\ 
			~ &   &[9]&\textbf{0.911} & \textbf{0.904} & \textbf{0.897} & 0.904 & 0.907 & \textbf{0.903}  \\ 
			\cline{2-3}
			~&\multirow{2}{*}{convolutional}&[11]&0.895 & 0.894 & 0.894 & 0.894 & 0.881 & 0.884  \\ 
			~&&[14]&0.902 & 0.897 & 0.889 & 0.906 & 0.891 & 0.887  \\ 
			\cline{2-3}
			~&\multirow{2}{*}{hybrid}&[16]&$\backslash$ & $\backslash$ & $\backslash$ & 0.932 & 0.902 & 0.866  \\
			~&&[18]&$\backslash$ & $\backslash$ & $\backslash$ & \textbf{0.943} & 0.935 & 0.902  \\ 
			\cline{2-3}
			~&other&[20]&0.799 & 0.766 & 0.739 & 0.795 & 0.769 & 0.752  \\ 
			\cline{1-3}
			L&hybrid&[25]&$\backslash$ & $\backslash$ & $\backslash$ & \textbf{0.943} & \textbf{0.936} & 0.883  \\ \hline
			
			\multicolumn{3}{c|}{Dataset} & \multicolumn{6}{c}{\textbf{Movie}}\\ 
			\hline
			\multicolumn{3}{c|}{bpw} & \multicolumn{3}{c|}{FLC} & \multicolumn{3}{c}{VLC}\\
			\cline{1-3}
			\multicolumn{2}{c|}{classification} & method & 1&2&3&1&2&3 \\ 
			\hline
			\multirow{9}{*}{S}  & \multirow{3}{*}{sequential} & [7] & 0.953&	0.923&	0.903&	0.947&	0.933&	0.917 \\ 
			~ & ~ & [8] &$\backslash$&$\backslash$&$\backslash$&0.977&	0.960&	0.918\\ 
			~ &   &[9]&0.962&	0.950&	0.932&	0.964&	0.948&	\textbf{0.937}  \\ 
			\cline{2-3}
			~&\multirow{2}{*}{convolutional}&[11]&0.949&	0.935&	0.918&	0.949&	0.931&	0.923  \\ 
			~&&[14]&0.954&	0.936&	0.921&	0.952&	0.939&	0.923  \\ 
			\cline{2-3}
			~&\multirow{2}{*}{hybrid}&[16]&$\backslash$ & $\backslash$ & $\backslash$ & 0.963&	0.946&	0.923  \\
			~&&[18]&$\backslash$ & $\backslash$ & $\backslash$ & 0.975&	0.960&	0.932  \\ 
			\cline{2-3}
			~&other&[20]&0.885&	0.833&	0.782&	0.878&	0.836&	0.802 \\ 
			\cline{1-3}
			L&hybrid&[25]&$\backslash$ & $\backslash$ & $\backslash$ & \textbf{0.970}&	\textbf{0.966}&	0.929 \\ \hline
			
		\end{tabular}
	\end{threeparttable}
	\begin{tablenotes}
		\item[] Note: Payloads: 1, 2, and 3bpw; datasets: Twitter and Movie. "S" and "L" represent "statistical vector embedding" and "language-model vector embedding". \textbf{Bold} represents the best performances. "$\backslash$" represents this method has no effect on the corresponding datasets.
	\end{tablenotes}
\end{table}

In terms of evaluation metrics, the detection performance is assessed using metrics such as Accuracy (Acc), Precision (P), Recall (R), and F1 score. The evaluation of computational resources includes metrics such as Time.

Since nearly ninety percent of existing work uses Twitter, Movie, and News as their raw data, RNN-Stega \cite{3} and Tina-Fang \cite{2} are used as the steganography schemes. The detection performance and time cost are shown in Table \ref{tab2} to Table \ref{tab4}. 

\begin{table}[!htbp]
	\centering
	\setlength{\tabcolsep}{0.7mm}
	\scriptsize
	\begin{threeparttable}
		\caption{Acc comparison of different methods in traditional laboratory data. (Stegenography scheme: Tina-Fang [2])}
		\label{tab3}
		\begin{tabular}{c|c|c|ccc|ccc|ccc}
			\hline
			\multicolumn{3}{c|}{Dataset} & \multicolumn{3}{c}{\textbf{Twitter}} & \multicolumn{3}{|c}{\textbf{Movie}} & \multicolumn{3}{|c}{\textbf{News}} \\ 
			\hline
			\multicolumn{3}{c|}{bpw} & \multirow{2}{*}{1} & \multirow{2}{*}{2} & \multirow{2}{*}{3} & \multirow{2}{*}{1} & \multirow{2}{*}{2} & \multirow{2}{*}{3} & \multirow{2}{*}{1} & \multirow{2}{*}{2} & \multirow{2}{*}{3} \\
			\cline{1-3}
			\multicolumn{2}{c|}{classification} & method & ~ & ~ & ~ & ~ & ~ & ~ & ~ & ~ & ~ \\ 
			\hline
			\multirow{10}{*}{S}  & \multirow{2}{*}{\shortstack[c]{sequ-\\ential}} & [7] & 0.791 & 0.850 & 0.924 & 0.910 & 0.963 & 0.973 & 0.915 & 0.924 & 0.968 \\ 
			~ & ~ & [8] & 0.783 & 0.842 & 0.912 & 0.906 & 0.958 & 0.970 & 0.917 & 0.923 & 0.972 \\ 
			\cline{2-3}
			~ & \multirow{3}{*}{\shortstack[c]{convol-\\utional}} & [9] & 0.749 & 0.846 & 0.921 & 0.943 & 0.964 & 0.985 & $\backslash$ & $\backslash$ & $\backslash$ \\ 
			& ~ & [12] & 0.780 & 0.826 & 0.920 & 0.878 & 0.950 & 0.961 & 0.904 & 0.896 & 0.965 \\ 
			& ~ & [14] & 0.835 & 0.929 & 0.962 & 0.859 & 0.939 & 0.967 & $\backslash$ & $\backslash$ & $\backslash$ \\ 
			\cline{2-3}
			~ & \multirow{4}{*}{hybrid} & [16] & 0.741 & 0.852 & 0.935 & \textbf{0.957} & \textbf{0.966} & \textbf{0.989} & $\backslash$ & $\backslash$ & $\backslash$ \\ 
			~ & ~ & [17] & 0.786 & 0.834 & 0.908 & 0.901 & 0.957 & 0.966 & 0.913 & 0.920 & 0.962 \\ 
			~ & ~ & [18] & \textbf{0.852} & 0.897 & 0.934 & 0.869 & 0.918 & 0.957 & 0.893 & 0.934 & 0.962 \\ 
			~ & ~ & [19] & 0.801 & 0.924 & 0.976 & 0.812 & 0.888 & 0.915 & $\backslash$ & $\backslash$ & $\backslash$ \\ 
			\cline{2-3}
			&other & [20] & 0.745 & 0.793 & 0.879 & 0.845 & 0.918 & 0.941 & 0.858 & 0.864 & 0.920 \\ 
			\cline{1-3}
			\multirow{2}{*}{L} & hybrid & [25] & 0.786 & 0.941 & \textbf{0.986} & $\backslash$ & $\backslash$ & $\backslash$ & \textbf{0.972} & 0.986 & 0.992 \\
			\cline{2-3}
			~ & other & [29] & 0.800 & \textbf{0.955} & 0.985 & 0.861 & 0.958 & 0.976 & \textbf{0.972} & \textbf{0.988} & \textbf{0.994} \\ \hline
		\end{tabular}
	\end{threeparttable}
	\begin{tablenotes}
		\item[] Note: Payloads: 1, 2, and 3bpw; datasets: Twitter, Movie and News. "S" and "L" represent "statistical vector embedding" and "language-model vector embedding". \textbf{Bold} represents the best performances. "$\backslash$" represents this method has no effect on the corresponding datasets.
	\end{tablenotes}
\end{table}

\begin{table}[!htbp]
	\centering
	\footnotesize
	\caption{Time comparison of different methods in traditional laboratory data scenarios (unit: seconds).}
	\label{tab4}
	\begin{tabular}{c|c|cc|cc}
		\hline
		\multicolumn{2}{c|}{Dataset} & \multicolumn{2}{c|}{\textbf{Twitter}} & \multicolumn{2}{c}{\textbf{Movie}} \\ 
		\hline
		classification & method & [2] & [3] & [2] & [3] \\ 
		\hline
		\multirow{3}{*}{S} & [7] & \textbf{1.22} & \textbf{1.31} & 2.77 & \textbf{2.84} \\ 
		\cline{2-2}
		~ & [11] & 1.27 & 1.33 & \textbf{2.57} & 2.92 \\ 
		\cline{2-2}
		~ & [16] & 4.38 & 4.47 & 5.71 & 6.02 \\ 
		\cline{1-2}
		L & [27] & 53.62 & 55.76 & 53.63 & 55.78 \\ 
		\hline
	\end{tabular}
\end{table}

By observing Table \ref{tab2} to Table \ref{tab4}, it can be found that the performances of the existing methods are: 1. The detection performance of the "language-model vector embedding" methods are generally better than that of the "statistical vector embedding" methods. 2. In a vector mapping, the detection performance of the "hybrid" methods is better than that of a single architecture ("sequential" and "convolutional"). 3. The training spatio temporal consumption of the "language-model vector embedding" methods are significantly higher than that of the "statistical vector embedding" methods.

\section{Conclusion and Outlook}\label{sec5}
So far, researchers have designed a series of deep-learning methods and frameworks, achieving excellent detection results, and the field has achieved unprecedented development. In this paper, the conclusions are as follows:

\begin{itemize}
	\item \textbf{Works classification}: Two-level classification of existing work. From the vector space mapping level, the existing work can be divided into two categories. From the feature extraction model level, these works can be further divided into four categories.
	
	\item \textbf{Learning paradigm}: Transductive learning solves the problem of domain mismatch for linguistic steganalysis; Meta learning captures the same points of different stego texts, enabling the network to be trained on detection with fewer samples; Transfer learning provides small-scale detection networks with a stronger ability to capture features. Lifelong learning reduces the degree of forgetting detected stego texts.
\end{itemize}

Through the conclusion, we also found some challenges and triggered thinking on the key issues to be solved:

\begin{itemize}
	\item \textbf{New learning paradigms can be introduced}. The existing work using the learning paradigm is well adapted to specific tasks and has obtained excellent detection results, so new learning paradigms are expected to bring corresponding solutions to new research motivations.
	
	\item \textbf{Over-reliance in the NLP needs to be shaken off}. Much of the current work is inspired by advanced natural language processing technologies, ignoring the nature of carrier or distribution changes caused by steganography embeddings. Researchers can design a highly interpretable steganalysis methods.
\end{itemize}

\end{document}